\documentclass[10pt,twocolumn,letterpaper]{article}

\usepackage{cvpr}              

\usepackage[dvipsnames]{xcolor}

\usepackage{graphicx}
\usepackage{subcaption}
\usepackage{bm}
\usepackage{booktabs}
\usepackage{siunitx}
\usepackage{colortbl, color}

\usepackage{comment}

\definecolor{cvprblue}{rgb}{0.21,0.49,0.74}
\definecolor{lblue}{rgb}{0.9,0.95,1}
\definecolor{lpurple}{rgb}{0.35,0.25,0.55}
\definecolor{lgreen}{rgb}{0.95,1,0.95}
\definecolor{sblue}{rgb}{0,0.45,1}
\definecolor{lgray}{gray}{0.95}
\definecolor{lyellow}{rgb}{1,1,0.92}

\usepackage[accsupp]{axessibility}

\usepackage[pagebackref,breaklinks,colorlinks,citecolor=cvprblue]{hyperref}

\def\paperID{*****} 
\def\confName{CVPR}
\def\confYear{2024}

\title{AIS 2024 Challenge on Video Quality Assessment of User-Generated Content: Methods and Results}

\author{
Marcos V. Conde~$^{*\dagger}$ \and
Saman Zadtootaghaj~$^{*}$ \and
Nabajeet  Barman~$^{*}$ \and
Radu Timofte~$^{*}$ \and
Chenlong He \and
Qi Zheng \and
Ruoxi Zhu \and
Zhengzhong Tu \and
Haiqiang Wang \and
Xiangguang Chen \and
Wenhui Meng \and
Xiang Pan \and
Huiying Shi \and
Han Zhu \and
Xiaozhong Xu \and
Lei Sun \and
Zhenzhong Chen \and
Shan Liu \and
Zicheng Zhang \and
Haoning Wu \and
Yingjie Zhou \and
Chunyi Li \and
Xiaohong Liu \and
Weisi Lin \and
Guangtao Zhai \and
Wei Sun \and
Yuqin Cao \and
Yanwei Jiang \and
Jun Jia \and
Zhichao Zhang \and
Zijian Chen \and
Weixia Zhang \and
Xiongkuo Min \and
Steve Göring \and
Zihao Qi \and
Chen Feng \and
}

\begin{document}

\maketitle

\let\thefootnote\relax\footnotetext{$*$ Marcos V. Conde ($\dagger$ corresponding author, project lead) and Radu Timofte are the challenge organizers, while the other authors participated in the challenge and survey. \\
\noindent Marcos V. Conde and Radu Timofte are with University of W\"urzburg, CAIDAS \& IFI, Computer Vision Lab.\\
Saman Zadtootaghaj, Marcos V. Conde and Nabajeet Barman are with Sony Interactive Entertainment, FTG.\\
AIS 2024 webpage:~\url{https://ai4streaming-workshop.github.io/}.
Code:~\url{https://github.com/mv-lab/VideoAI-Speedrun}} 

\begin{abstract}
This paper reviews the AIS 2024 Video Quality Assessment (VQA) Challenge, focused on User-Generated Content (UGC). The aim of this challenge is to gather deep learning-based methods capable of estimating the perceptual quality of UGC videos. The user-generated videos from the YouTube UGC Dataset include diverse content (sports, games, lyrics, anime, etc.), quality and resolutions. The proposed methods must process 30 FHD frames under 1 second. In the challenge, a total of 102 participants registered, and 15 submitted code and models. The performance of the top-5 submissions is reviewed and provided here as a survey of diverse deep models for efficient video quality assessment of user-generated content.
\end{abstract}

\setlength{\abovedisplayskip}{1pt}
\setlength{\belowdisplayskip}{1pt}

\section{Introduction}

\begin{figure}[t]
    \centering
    \includegraphics[width=\linewidth]{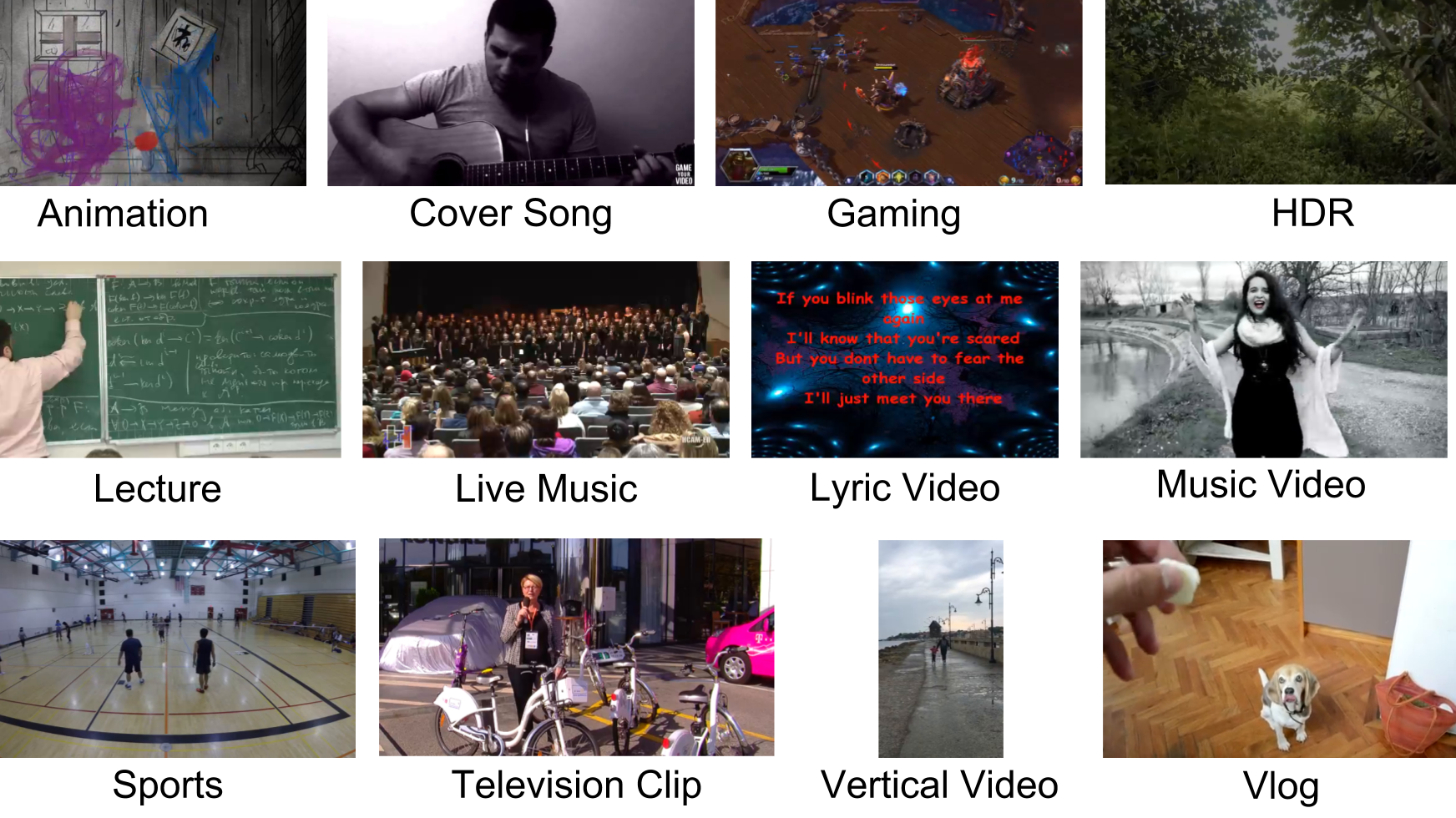}
    \caption{Samples from the videos in the YT-UGC Dataset~\cite{wang2019youtube}.}
    \label{fig:teaser}
\end{figure}

Past two decades have seen a massive increase in popularity and demand for online video streaming applications such as Netflix and YouTube~\cite{statOTT}. This has been made possible due to improvements in network capacity, improved end-user devices, and increased computational efficiency, allowing users to stream and watch content for hours over the internet everyday~\cite{statpermin}. In order to optimize the end-user experience and provide them with an improved quality of experience, the service provider must measure the perceptual quality of the videos being delivered to them.

Image quality assessment (IQA) or video quality assessment (VQA) can be assessed either subjectively or objectively. In subjective quality assessments, the users directly assess the image/video quality and provide a rating for that~\cite{gu2022ntire, sinno2018large, hosu2017konstanz, wang2019youtube}. However, such assessment processes are time consuming, costly, and often not realistic in real-world applications. Objective quality models help bridge this gap by using mathematical/statistical models to predict the quality as would be subjectively judged by human observers~\cite{vmaf}. In recent years, deep learning techniques have enabled us to learn objective quality metrics from visual content and the corresponding ratings. Depending on the availability of a reference, QA models can broadly be classified into Full-Reference and No-Reference (Blind)~\cite{Barman2019Survey}.


This challenge deals with the design of deep learning-based methods for blind video quality metrics, targeting user-generated content. Given a short video of an arbitrary resolution, the method will predict the overall quality.

In this context, user-generated content refers to content that is captured by users using consumer-grade devices, such as (primarily) smartphones, tablets, GoPros, etc. (see \cref{fig:teaser}), and often shared via platforms such as Instagram, YouTube, TikTok, etc~\cite{wang2019youtube, wang2021rich, sinno2018large}. Unlike professionally generated content, they are usually captured under very challenging conditions, and hence, these can suffer from many artifacts (camera capture impairments, lightning conditions, formats (resolution, fps), etc.). 


Recent works focus on designing NR models using deep learning approaches on large-scale datasets to tackle this problem~\cite{wang2021rich,wu2023dover,wen2024modular}. Deep learning methods are better able to capture and model various factors such as content, distortion, compression, and blur artifacts while also taking into account the temporal aspect for video quality prediction. However, these demand large amounts of annotated data, this has led to the creation of larger, more realistic datasets such as KonVid-1k~\cite{hosu2017konstanz}, YouTube YT-UGC~\cite{wang2019youtube}, and more recently, KonViD-150k VQA Database~\cite{hahn2021}.

\section{UGC Video Quality Challenge}

\subsection{Dataset}

The challenge uses the YouTube User-Generated-Content (YT-UGC) dataset \cite{wang2019youtube} that consists of around 1000 video clips with a duration of 20 seconds. 

The dataset includes several perceptual artifacts such as blockiness, blur, banding, noise, and jerkiness. In addition, the dataset has a wide range of content types with 15 distinct categories, including animation, gaming, cover songs, music videos, and vlogs, among others. 

Moreover, a wide range of resolutions is considered in the dataset, including 360p, 480p, 720p, 1080p, and 2160p. 

The clips are annotated with subjective ratings in the 5-point categorical Absolute Rating (ACR) Scale. All videos were rated by more than 100 subjects using crowdsourcing. The Mean Opinion Score (MOS) is obtained on a rating scale of 1 to 5, where 1 is the lowest perceived quality (bad) and 5 is the highest perceived quality (excellent). 

For this AIS UGC Video Quality Assessment Challenge, the dataset is split into two sets, training and test. The larger portion of the dataset consisting of 900 clips is used for training, while the test set includes 126 clips, selected carefully considering a balanced range of resolution, content type, and distortion. We show samples in \cref{fig:teaser}.

\subsection{Model Design Rules}


\begin{itemize}
   \item The VQA models should be able to process FHD and HD clips of 30 frames under 1 second. 
   Frame sampling is allowed, as long as the runtime per frame is still $\leq$ 33 ms. This was measured on an NVIDIA A100 GPU (or similar modern GPUs \eg RTX 3090 Ti). 
   
   \item We use standard correlation metrics of model scores with subjective (MOS) ratings (SRCC, PCC, KCC).
   
   \item Participants were allowed to use any pre-trained and existing solutions.

   \item The organizers validate the efficiency and reproducibility of the methods.
\end{itemize}

\begin{table*}[t]
    \centering
    \resizebox{\linewidth}{!}{
    \begin{tabular}{l l c c c c c c c c c c}
         \toprule
         \rowcolor{lgray} Team & Method & SROCC & KROCC & PLCC & \# Params.~[M] & \multicolumn{2}{c}{Runtime [ms]} & \multicolumn{2}{c}{MACs~[G]}  \\
         \rowcolor{lgray} & &  &  &  &  &  30-FHD & 60-HD & 30-FHD & 60-HD \\
         \toprule
         FudanVIP & COVER~\cite{He2024cover} & 0.914 & 0.741 & 0.912 & 61.02 & 79.37 & 78.66 & NA & NA \\

         TVQE & TVQE & 0.915 & 0.741 & 0.918 & 8254 & 299.18 & 294.93 & 1127.35 & 1263.53 \\
         
         Q-Align & Q-Align~\cite{wu2023q} & 0.908 & 0.734 & 0.912 & 8198 & 526.55 & 429.4 & 991.17 & 991.17 \\
         
         SJTU MMLab & SimpleVQA+~\cite{sun2023analysis} & 0.906 & 0.728 & 0.911 & 207.7 & 222.96 & 394.51 & 140.17 & 280.35 \\
         
         AVT & AVT & 0.877 & 0.690 & 0.878 & 168 & 90.57 & 81.90 & NA & NA \\
         
         BVI-VQA & FasterVQA~\cite{wu2023neighbourhood} & 0.817 & 0.638 & 0.751 & 28.13 & 52.49 & 55.87 & NA & NA \\
         
         \midrule
         \rowcolor{lblue} Baseline & NDNet~\cite{ndnetgaming}  & 0.718 & 0.502 & 0.715 & 6.95 & 52.95 & 24.21 & 597.47 & 265.99 \\
         \rowcolor{lblue} Baseline & MobNet & NA & NA & NA & 2.22 & 157.74 & 138.65 & 397.31 & 353.60 \\
         \bottomrule
    \end{tabular}}
    \caption{\textbf{AIS 2024 UGC Video Quality Assessment Challenge Benchmark.} We report runtime and MACs operations for a complete 30-frame FHD clip, and 60-frame HD clip. ``NA" indicates the results are not available or could not be calculated.}
    \label{tab:efficiency}
\end{table*}

\begin{table}[t]
    \centering
    \resizebox{\linewidth}{!}{
    \begin{tabular}{r c c c c}
         \toprule
         \rowcolor{lgray} Method & SROCC & KROCC & PLCC & RMSE  \\
         \toprule
         BRISQUE~\cite{mittal2012no}        & 0.4398 & 0.2934 & 0.4525 & 0.5608  \\
         GM-LOG~\cite{xue2014blind}         & 0.3501 & 0.2336 & 0.3424 & 0.5904  \\
         VIDEVAL~\cite{9405420}             & 0.7946 & 0.5959 & 0.7691 & 0.4024 \\
         RAPIQUE~\cite{tu2021rapique}       & 0.7483 & 0.5556 & 0.7482 & 0.4177 \\
         FAVER~\cite{zheng2024faver}        & 0.7897 & 0.5832 & 0.7898 & 0.3861 \\
         NIQE~\cite{6353522}                & 0.2479 & 0.1689 & 0.3146 & 0.5976 \\
         HIGRADE~\cite{kundu2017no}         & 0.7639 & 0.5524 & 0.7507 & 0.4156 \\
         FRIQUEE~\cite{FRIQUEE}             & 0.7182 & 0.5268 & 0.7091 & 0.4439 \\
         CORNIA~\cite{6247789}              & 0.5988 & 0.4113 & 0.5905 & 0.5064 \\
         TLVQM~\cite{korhonen2019two}       & 0.6690 & 0.4833 & 0.6412 & 0.4831 \\
         CLIPIQA+~\cite{wang2023exploring}  & 0.5374 & 0.3734 & 0.5801 & 0.5128 \\
         FasterVQA~\cite{wu2023neighbourhood} & 0.5345 & 0.3716 & 0.5438 & 0.5284 \\
         FASTVQA~\cite{wu2022fast}          & 0.6493 & 0.4676 & 0.6792 & 0.4621 \\
         DOVER~\cite{wu2023dover}                 & 0.7359 & 0.5391 & 0.7653 & 0.4053 \\
         FasterVQA*                         & 0.6937 & 0.4965 & 0.6909 & 0.4552 \\
         FASTVQA*                           & 0.8617 & 0.6716 & 0.8669 & 0.3139 \\
         DOVER*                             & 0.8761 & 0.6865 & 0.8753 & 0.3144 \\
         \midrule

         FasterVQA*~(\cref{sec:bvi}) & 0.8170 & 0.6380 & 0.7510 & - \\
         
         AVT (\cref{sec:AVT})                  & 0.8775  & 0.6909  & 0.8785  & - \\

         SimpleVQA+~\cite{sun2023analysis} & 0.9060 & 0.7280 & 0.9110 & - \\
         
         Q-Align~\cite{wu2023q} & 0.9080 & 0.7340 & 0.9120 & - \\
         
         TVQE (\cref{sec:tvqe}) &  0.9150 &  0.7410 &  0.9182  & - \\
         
         COVER~\cite{He2024cover}  & {0.9143} & {0.7413} & {0.9122} & {0.2519} \\
         \bottomrule
    \end{tabular}
    }
    \caption{Extended comparison with classical and previous \emph{state-of-the-art} methods. We report some numbers from~\cite{He2024cover}. ``*" indicates models were fine-tuned using the AIS Challenge dataset.
    }
    \label{tab:results}
\end{table}

\begin{table}[t]
    \centering
    \resizebox{\linewidth}{!}{
    \begin{tabular}{l l c c c c c}
         \toprule
         \rowcolor{lgray} Team & Method & \# Params. & Runtime & MACs  \\
         \rowcolor{lgray} & & [M] & [ms] & [G] \\
         \toprule

         FudanVIP & COVER~\cite{He2024cover} & 61.02 & 79.37 & NA \\

         TVQE & TVQE & 8254 & 705.30 & 1127.35  \\
         
         Q-Align & Q-Align~\cite{wu2023q} & 8198 & 1707.06 & 991.17  \\
         
         SJTU MMLab & SimpleVQA+~\cite{sun2023analysis} & 207.7 & 245.512 & 140.175 \\
         
         \midrule
         \rowcolor{lblue} Baseline & NDNet~\cite{ndnetgaming}  & 6.95 & 209.43 & 479.06 \\
         \rowcolor{lblue} Baseline & MobNet & 2.22 & 347.51 & 1585.32 \\
         \bottomrule
    \end{tabular}
    }
    \caption{\textbf{High-Resolution Efficiency study} using as input a clip of 30 frames of 4K resolution $3840\times2160$. We report the runtime and MACs for the complete clip of 30 frames.}
    \label{tab:hr_efficiency}
\end{table}

\section{Challenge Results}

\subsection{Baselines}

We consider two Baseline models for benchmarking which are discussed next. 

NDNetGaming \cite{ndnetgaming} is a CNN-based quality metric that is designed to assess gaming video quality. NDNetGaming is designed to predict quality in an interpretable range of one to five, where one is the lowest quality, and five is the highest quality score. NDNetGaming uses DenseNet-121 as the backbone and is pre-trained on a large-scale gaming video dataset annotated with VMAF and fine-tuned by a public gaming video dataset. 
Since NDNetGaming was tailored for images, we used a sampling rate of 5 frames per second and averaged the resultant quality estimation. 

We additionally used MobileNet v2 as the second baseline model, which allows us to compare the efficiency of proposed models with a lightweight CNN image encoder architecture. We first process each frame using MobileNet~\cite{sandler2018mobilenetv2}. Next, we average the encoded features for all the frames obtaining a single deep encoded representation, and finally, we predict the quality using a single linear layer. Thus, no frame sampling is applied to the MobileNet result. This represents a naive solution for benchmarking purposes. The baselines are highlighted in blue in \cref{tab:efficiency}.

\subsection{Architectures and main ideas}


\begin{enumerate}
    \item \textbf{Frame Sampling:} Given a clip of $N$ frames, most methods apply temporal (down)sampling \ie process 1 (or 2) frames of every 30. This allows to increase efficiency without harming performance. Note that this is the reason why we report clip-based metrics instead of frame-based metrics. For instance, a model can virtually process a 30-frame clip in 100 ms, yet it does not imply a 330 FPS performance.

    \item \textbf{Spatial Downsampling:} Besides pooling in the temporal domain, most approaches resize the frames to lower resolutions (\eg 512px) to reduce memory requirements and operations.

    \item \textbf{Ensembles:} The best solutions such as COVER~\cite{He2024cover} and TVQE use multiple image processing models to extract diverse features~\cite{wang2021rich}. Each model is trained to focus on predicting specific properties such as aesthetics or compression. Although combining multiple models might increase training and inference complexity, this approach provides the best performance while being surprisingly efficient.
    
\end{enumerate}

\subsection{Efficiency Study}

In \cref{tab:efficiency} we present the summary of quantitative results and efficiency metrics for each method. The efficiency metrics are calculated using: \url{https://github.com/mv-lab/VideoAI-Speedrun}. The runtime is the average of 10 independent runs (after GPU warmup).

TVQE and Q-Align~\cite{wu2023q} use novel LLM-based VQA approaches, thus the number of parameters is considerably high (8 Billion). These approaches leverage video descriptions and visual information to provide accurate ratings. Although the number of parameters and operations is considerably high, the models can process 30 frames under a second, even at high resolution (FHD, 4K).

As we show in \cref{tab:efficiency} and \cref{tab:hr_efficiency}, all the proposed methods can process 30 FHD frames in under 1 second, and 60 HD frames in under 0.5 seconds. Moreover, most approaches can process 30 4K frames under 1 second. 

\vspace{-3mm}

\paragraph{Discussion on frame-wise metrics}
We report clip-based metrics. Since each method uses different frame sampling techniques, it is difficult to calculate FPS or frame-wise metrics. We instead focus on 30-frame and 60-frame clips.

We can appreciate in \cref{tab:efficiency} that COVER~\cite{He2024cover}, TVQE and Q-Align~\cite{wu2023q} have almost constant runtime (or operations) independently of the input resolution or number of frames. The reason is the constant temporal-spatial
downsampling on the input video \ie FHD, HD, and 4K frames are 
always downsampled to the same resolution and fed into the model. 


\paragraph{Related Challenges}
This challenge is one of the AIS 2024 Workshop associated challenges on: Event-based Eye-Tracking~\cite{wang2024ais_event}, Video Quality Assessment of user-generated content~\cite{conde2024ais_vqa}, Real-time compressed image super-resolution~\cite{conde2024ais_sr}, Mobile Video SR, and Depth Upscaling. 

\newpage

\section{Challenge Methods and Teams}
\label{sec:teams}

In the following sections we describe the best challenge solutions. Note that the method descriptions were provided by each team as their contribution to this survey. 

\subsection{A Comprehensive Video Quality Evaluator}
\label{sec:cover}


\begin{center}

\vspace{2mm}
\noindent\emph{\textbf{Team FudanVIP}}
\vspace{2mm}

\noindent\emph{Chenlong He~$^1$,
Qi Zheng~$^1$,
Ruoxi Zhu~$^1$,
Zhengzhong Tu~$^2$}

\vspace{2mm}

\noindent\emph{
$^1$ State Key Laboratory of Integrated Chips and Systems, Fudan University, China\\
$^2$ University of Texas at Austin, America
}

\vspace{2mm}

\noindent{\emph{Contact: \url{zhengzhong.tu@utexas.edu}}}

\end{center}

The team introduces COVER~\cite{He2024cover}, a comprehensive video quality evaluator, a novel framework designed to evaluate video quality holistically --- from a technical, aesthetic, and semantic perspective.
Specifically, COVER leverages three parallel branches: (1) a Swin Transformer~\cite{liu2021swin} backbone implemented on spatially sampled crops to predict technical quality; (2) a ConvNet~\cite{liu2022convnet} employed on subsampled frames to derive aesthetic quality; (3) a CLIP\cite{radford2021learning} image encoder executed on resized frames to obtain semantic quality.
We further propose a simplified cross-gating block to interact with the three branches before feeding into the predicting head.
The final quality score is attained using a weighted sum of each sub-score, making a multi-faceted, explainable metric.
Our experimental results demonstrate that COVER exceeds the state-of-the-art models in multiple UGC video quality datasets while it is capable of processing 1080p videos in real-time.

\begin{figure*}[htbp]
    \centering
    \includegraphics[width=\textwidth]{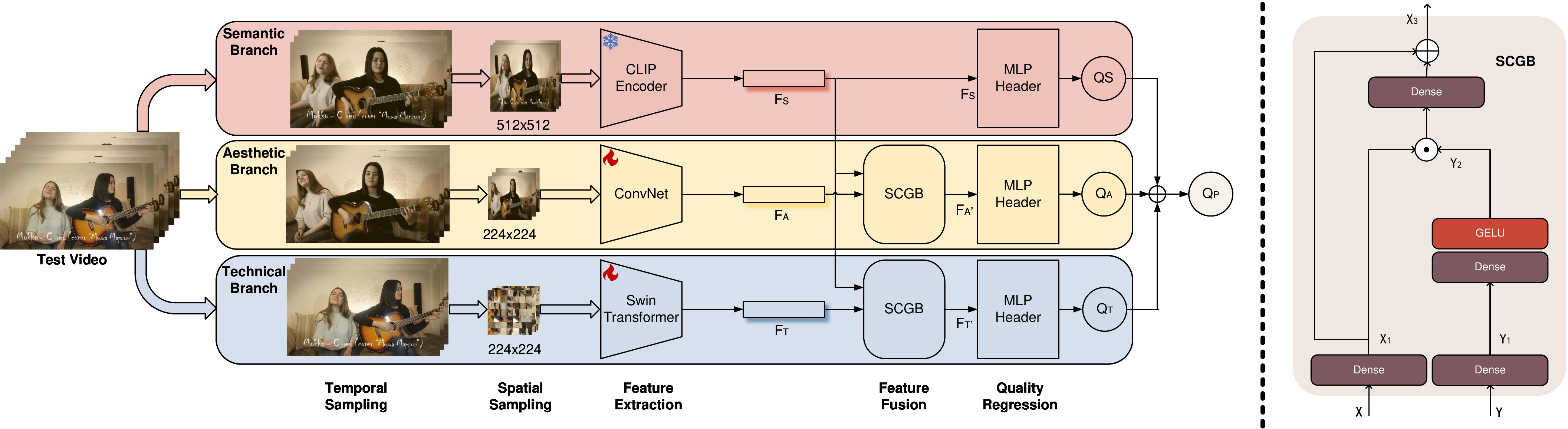}
    \caption{The architecture of our proposed \textbf{\underline{C}}\textbf{\underline{O}}mprehensive \textbf{\underline{V}}ideo quality \textbf{\underline{E}}valuato\textbf{\underline{R}} (\textbf{COVER}).
COVER processes a video clip in three parallel branches: 1) a semantic branch that extracts high-level object-semantics-related information using a pre-trained CLIP image Encoder; 2) an aesthetic branch that leverages a ConvNet run on subsampled image thumbnails to analyze their looking; 3) a technical branch utilizing Swin Transformer to execute on fragments. We also 
devise a simplified cross-gating block (SCGB) to fuse multi-branch features together, yielding the final quality score.
}
    \label{fig:my_diagram}
\end{figure*}

\subsubsection{Method}
The network architecture of our proposed \textbf{\underline{C}}\textbf{\underline{O}}mprehensive \textbf{\underline{V}}ideo quality \textbf{\underline{E}}valuato\textbf{\underline{R}} (\textbf{COVER}) is illustrated in Fig.\ref{fig:my_diagram}. This network accepts videos that have been subjected to temporal-spatial sampling as its input. Its architecture is divided into three branches: a CLIP-based semantic branch, an aesthetic branch and a technical branch, each consisting of a feature extraction module and a quality regression module. Notably, aesthetic and technical branches additionally incorporate a feature fusion module to integrate features from the semantic branch. The input video is processed through these branches to generate three scores, reflecting the video’s quality across the respective dimensions. The final score is the average of scores from three dimensions.

\subsubsection{Temporal and Spatial Sampling}
As shown in Fig. \ref{fig:my_diagram}, before serving as input to each branch's feature extraction module, the input videos first undergo temporal-spatio sampling. To enhance the real-time performance of the network, temporal sampling is designed to be very sparse. In the temporal sampling process for the input video, the semantic branch samples one frame out of every thirty frames, while the aesthetic and technical branches sample two frames out of every thirty frames.

For spatial sampling, the semantic and aesthetic branches resize the video resolution to 512x512 and 224x224, respectively. The technical branch, however, employs a fragment operation, where a frame from the video is divided into 7x7 sub-blocks. These sub-blocks are then randomly sampled and reassembled into a frame with a resolution of 224x224.
\subsubsection{Feature Extraction}
Several studies have demonstrated the effectiveness of CLIP~\cite{radford2021learning}, a foundation model, in both IQA~\cite{wang2023exploring} and VQA~\cite{wu2023dover} tasks. By extracting semantic information from images and videos, CLIP can accurately assess their subjective quality. However, the aforementioned studies did not address the more challenging task of UGC-VQA. This motivates us to employ the Image Encoder of CLIP as the backbone of the feature extraction module for the semantic branch. The pretrained weights (ViT-L/14) on OpenAI is imported and frozen.

For the technical branch, the Swin Transformer~\cite{liu2021swin} is utilized as the backbone of the feature extraction module. A CNN network, specifically the ConvNet~\cite{liu2022convnet}, is used as the backbone of the feature extraction module for the aesthetic branch. These two branches are initialized with weights pretrained on the LSVQ~\cite{ying2021patch} from DOVER~\cite{wu2023dover}, and it will be fine-tuned during subsequent training.
\subsubsection{Feature Fusion}
CLIP's image encoder is endowed with robust capabilities in representing image semantics by its numerous training samples. Thus, the abundant information contained in CLIP's output features may inherently correlate with the features of the other branches. To fully harness the representative features generated by the semantic branch and let it modulate the other branches, we propose a feature fusion block. More specifically, we modify the cross-gating block~\cite{tu2022maxim}, and name it Simple Cross-Gating Block (SCGB), for feature fusion between the semantic-aesthetic and semantic-technical feature pairs. As illustrated in Fig.~\ref{fig:my_diagram}, The fused features from the aesthetic and technical branches, along with the features from the semantic branch, are then fed into their respective quality regression modules.

The detailed architecture of SCGB is depicted in Fig.~\ref{fig:my_diagram}. The input of the block are two tensors $X$ and $Y$. $X$ is the feature from the technical or aethetic branch, while $Y$ is from the CLIP-based semantic branch. After the input channel projections are applied, the projected CLIP features are fed to a gating pathway to yield the gating weights, which are then multiplied by the features from the other branch. Finally, the output projection and residual connection are applied. 
\subsubsection{Quality Regression}
The features from each branch are individually fed into a multi-layer perceptron (MLP) Header to predict quality scores, i.e., $Q_S$, $Q_A$, and $Q_T$, as shown in Fig.~\ref{fig:my_diagram}, and the final predicted quality, $Q_P=(Q_S+Q_A+Q_T)/3$.
To enforce that each branch can independently capture the features of its focused dimension and accurately predict video quality, we adopted the limited view biased supervision scheme~\cite{wu2023dover}, which minimizes the relative loss between predictions in each branch with the overall opinion MOS, as formulated below:
\begin{equation}
\begin{split}
\mathcal{L}_{all}=&\mathcal{L}_{rel}(Q_{S},\text{MOS})+\mathcal{L}_{rel}(Q_{A},\text{MOS})\\
+&\mathcal{L}_{rel}(Q_{T},\text{MOS})
\label{eq:loss}
\end{split}
\end{equation}

\subsubsection{Inference Time}
VQA models are highly practical tools potentially deployed on large-scale video streaming platforms to process millions of video streams every day.
Therefore, the actual inference cost per video is highly significant to the system's total performance and revenue.
We have imbued efficient modular design in every aspect of the COVER model, leading to highly efficient inference speed.
We benchmarked the model inference time required by COVER on a video clip of 30 frames of 1080p resolution using a TITAN RTX graphic card. As shown in Table~\ref{tab:inference_time}, COVER's semantic, aesthetic, and technical branch demands 191, 96, and 23 milliseconds to complete, together adding up to a total inference time of 311 milliseconds.
In other words, this inference latency translates to a highly efficient VQA metric that attains state-of-the-art performance with explainable properties and inferences at \textbf{96 fps}, almost 3x faster than real-time processing speed.
\begin{table}[htbp]
  \centering
  \small
  \setlength{\tabcolsep}{8pt}
  \caption{Inference time of COVER on a 30-frame chunk of a 1080p video on a TITAN RTX GPU card. The total 311 ms inference time translates to \textbf{96 fps}, 3x faster than real-time processing.}
    \begin{tabular}{l|cccc}
    \toprule
    Branch & Semantic & Aesthetic & Technical & All \\
    \toprule
    Time (ms) & 191 & 96 & 23 & 311 \\
    \bottomrule
    \end{tabular}%
  \label{tab:inference_time}%
\end{table}%

\paragraph{Implementation details}

The hyper-parameter settings within COVER for its various components are outlined as follows: i) the backbone of the feature extraction module for semantic branch is the Image Encoder from CLIP~\cite{radford2021learning} of type ViT-L/14; ii) the feature extraction backbone of aesthetic branch is a ConvNet~\cite{liu2022convnet}, structured into 4 stages. The configuration of each stage, defined by the number of blocks and feature dimensions, is as follows: (3, 96), (192, 3), (384, 9), and (768, 3); iii) the feature extraction backbone of technical branch is a Swin Transformer~\cite{liu2021swin}, which also comprises 4 stages. Within each stage, the number of heads is set to 3, 6, 12, and 24, respectively, with the number of projection output channels being 96; iv) the SCGB module operates with input and output feature dimensions both set to 768, and its dropout layer has a drop ratio of 0.1; v) the input feature dimension for the MLP Header module is 768. It includes two dropout layers, both with a drop ratio of 0.5.

The training process for our model is structured into three distinct stages:
\begin{enumerate}
    \item \textbf{Initial Training of Technical and Aesthetic Branches:}
    Initially, we train the entire network for both the technical and aesthetic branches. During this stage, the weights of both backbones and MLP Headers for all branches are fine-tuned. 
    \item \textbf{Integrating Semantic Branch and Further Fine-tuning:}
    Building on the best weights obtained from stage 1, we integrate the semantic branch into model. Then MLP Headers of all branches, along with backbones of both technical and aesthetic branches are fine-tuned.
    \item \textbf{Incorporation of SCGB and Final Fine-tuning:}
    Based on the optimal weights from stage 2, we add two SCGBs to model. Subsequent fine-tuning of both SCGBs along with all MLP Headers is conducted.
\end{enumerate}

Given the specific validation set of YouTube-UGC, our multi-stage training approach maintains the same data split across each step, allowing for incremental improvements in training effectiveness. 

Throughout different training stages, only the specific training set of YouTube-UGC is used. For training strategies. we employ the ADAM optimizer with an initial learning rate of $1 \times 10^{-3}$ and a cosine learning rate decay strategy with a decay weight of 0.05, over a total of 20 epochs. However, the batch size varies across different stages, being set to 10, 8, and 24 respectively. Our network, implemented in the Pytorch framework and running on an A6000 GPU card, approximately requires one day to complete the entire training process.

\begin{figure*}[t]
\centering
\includegraphics[width=\textwidth]{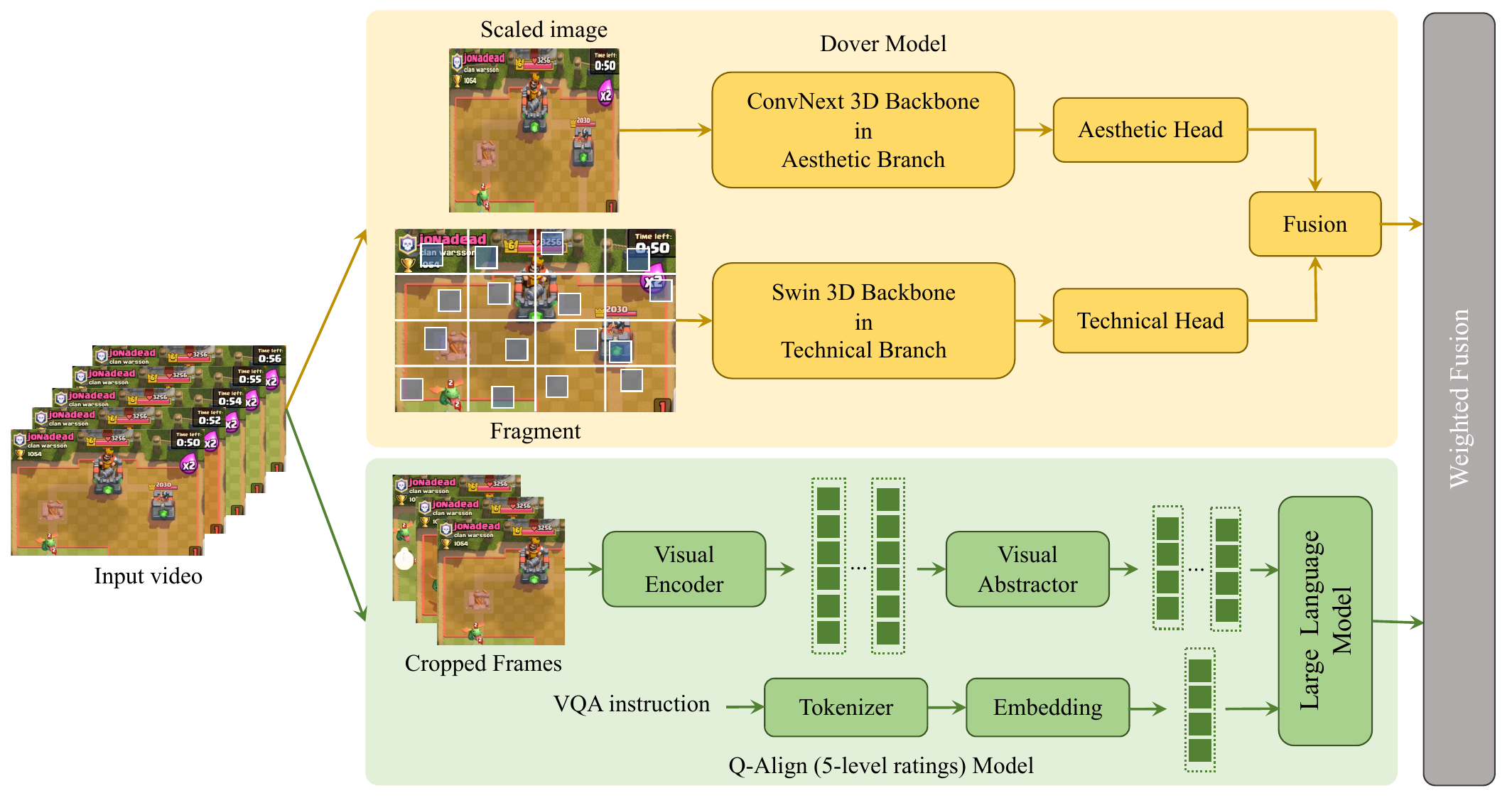} 
\caption{The architecture of the TVQE method.}
\label{fig:framework}
\end{figure*}

\subsection{TVQE: Tencent Video Quality Evaluator}
\label{sec:tvqe}


\begin{center}

\vspace{2mm}
\noindent\emph{\textbf{Team TVQE}}
\vspace{2mm}

\noindent\emph{
Haiqiang Wang~$^1$,
Xiangguang Chen~$^1$,
Wenhui Meng~$^1$,
Xiang Pan~$^1$,
Huiying Shi~$^2$,
Han Zhu~$^2$,
Xiaozhong Xu~$^1$,
Lei Sun~$^1$,
Zhenzhong Chen~$^2$,
Shan Liu~$^1$
}

\vspace{2mm}

\noindent\emph{
$^1$ Tencent\\
$^2$ Wuhan University
}

\end{center}

TVQE is a hybrid model trained for VQA tasks. The proposed method fully takes into account several aspects of video quality subjective assessment: 1. Humans make judgments with attention to both global semantic and local visual information; 2. Subjective evaluation experiments usually require observers to learn and judge in discrete text-defined levels. 
Therefore, it combines three networks, \emph{i.e.}, IQA network, DOVER~\cite{wu2023dover}, and Q-Align~\cite{wu2023q} model, to extract visual information and semantic information and predicts the subjective quality more accurately via weighted fusion operation. The framework of the proposed method is shown in Fig.\ref{fig:framework}.

First, considering that humans have a strong perception of visual information in the spatial dimension when making the judgment, we introduce a feature pyramid aggregation mechanism on the backbone, \emph{i.e.}, the ConvNeXt, to extract visual representations of the key frame. The pyramid structure facilitates the full utilization of the extracted information as well as better exploitation of the shallow visual features. 
Then, considering the influence of video content on subjective assessment, we use the DOVER model~\cite{wu2023dover} with 3D convolution to assess video quality through aesthetic and technical branches.

\begin{table}[!t]
    \centering
 	\resizebox{\linewidth}{!}{
        \begin{tabular}{l c c c}		
            \toprule
            Variant & Fusion Ratio & SROCC  & PLCC \\
            \midrule
            DOVER (v0) &-&0.822 &0.830  \\
            DOVER (v1) &-&0.881 &0.887  \\
            \midrule
            Q-Align5 (v0)&-& 0.842 &0.838  \\
            Q-Align5 (v1)&-& 0.895 &0.885  \\
            Q-Align5 (v2)&-& 0.908 &0.871  \\
            \midrule
            DOVER+Q-Align5 & 7:8 &0.913  &0.915 \\
            \bottomrule
        \end{tabular}
        }
        \caption{Performance of Different TVQE Variants. DOVER (v0) represents the pre-trained model, and (v1) the fine-tuned model. Q-Align5 (v0) represents the pre-trained model, (v1) represents the results by finetuning Visual Abstractor, and (v2) represents the results by finetuning the last 5 transformer layers in Visual Encoder and Visual Abstractor.
        } 
	\label{tab:ablation}
\end{table}


Finally, we adopt a large multi-modality model, \emph{i.e.}, Q-align~\cite{wu2023q}, to fit the fact that subjective judgment is usually in discrete text-defined levels. The purpose is to stimulate the behavior of the human annotation process by tuning LLMs (Large Language Models).

These three models were trained independently on the official YT-UGC dataset~\cite{wang2019youtube} following the challenge splits. During the inference stage, the final predicted score could be obtained by heuristically fusing the prediction results of these models.

\paragraph{Ablation Study} Table~\ref{tab:ablation} gives the ablation study of submitted solution. We finetuned the SOTA DOVER and Q-align model on the give YT-UGC dataset. We take a small split from the training set as the second validation set for model selection. 

For the DOVER architecture, it could be seen that the SROCC value increases from 0.822 to 0.881 after carefully finetuning parts of the original network. 
For the Q-align architecture, we tried different finetune strategy. Empirically, we found that finetuning the last 5 layers of the visual encoder and the visual abstractor block gives the best performance gain, \emph{i.e.}, 0.07 in terms of SROCC. 

Then, thanks to the ensemble strategy, the performance is further boosted by 0.005 in terms of SROCC and 0.44 in terms of PLCC, respectively. 

\paragraph{Inference} The processing time for 30 frames with 4K resolution on the NVIDIA RTX 3090 GPU is 0.8 seconds, which meets the required 30 FPS. Thus, the inference runtime with other lower resolutions (e.g., 2K and 1K resolutions) can also satisfy the 30-frames under 1s requirement.

\paragraph{Implementation details}

\begin{itemize}
    \item \textbf{Framework:} Pytorch (version 2.0.1)
    \item \textbf{Optimizer and Learning Rate:} AdamW with initial learning rate 2e-5
    \item \textbf{GPU:} NVIDIA Tesla A100 (40G)
    \item \textbf{Datasets:} YT-UGC dataset (challenge split)
    \item \textbf{Training Time:} approximately 10 hours.
    \item \textbf{Training Strategies:} Initialization with the public pre-trained model, and training for several epochs.
\end{itemize}

\begin{figure*}[!ht]
    \centering
    \includegraphics[width=\linewidth]{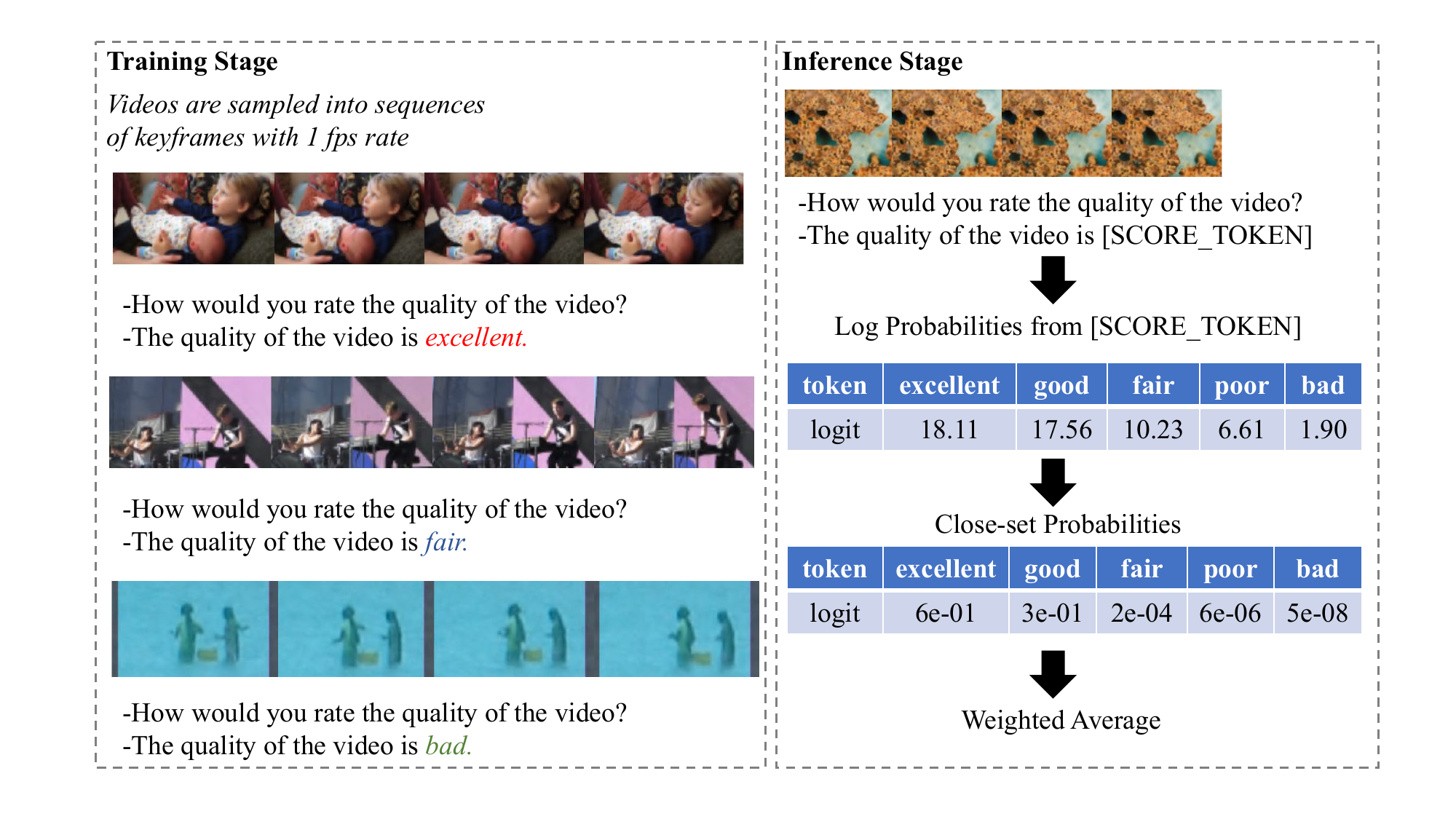}
    \caption{The framework of Q-Align~\cite{wu2023q}, where we feed quality question-answer pairs to train LMMs and obtain the 5-level quality probabilities during the inference stage.}
    \label{fig:qalign}
\end{figure*}

\subsection{Q-Align: Aligning video quality with text descriptions based on LMM}
\label{sec:qalign}


\begin{center}

\vspace{2mm}
\noindent\emph{\textbf{Team Q-Align}}
\vspace{2mm}

\noindent\emph{Zicheng Zhang~$^1$,
Haoning Wu~$^2$,
Yingjie Zhou~$^1$,
Chunyi Li~$^1$,
Xiaohong Liu~$^1$,
Weisi Lin~$^2$,
Guangtao Zhai~$^1$}

\vspace{2mm}

\noindent\emph{$^1$ Shanghai Jiao Tong University\\
$^2$ Nanyang Technological University}

\vspace{2mm}

\noindent{\emph{Contact: \url{zzc1998@sjtu.edu.cn}}}

\end{center}

We convert the traditional mean opinion scores (MOS) and the corresponding video into question-answer pairs to teach LMM VQA knowledge. Then we acquire the probabilities of the video quality from LMM response and obtain the final quality values via weighted average.

Q-Align~\cite{wu2023q} is based on large multi-modality models (LMMs). During the training stage, we divide the quality labels into specific rating categories. Given that the human-assigned ratings are evenly spaced, we utilize equally spaced intervals for transforming scores into these categories. We achieve this by evenly dividing the range from the maximum score ($\mathrm{M}$) to the minimum score ($\mathrm{m}$) into five separate intervals, assigning scores within each interval to corresponding categories:
\begin{equation}
{L(s)} = l_i \text{ if } \mathrm{m} + \frac{i-1}{5} \times (\mathrm{M} - \mathrm{m}) < s \leq \mathrm{m} + \frac{i}{5} \times (\mathrm{M} - \mathrm{m})
\end{equation}
where the set {$l_i|_{i=1}^{5}$} = \{\textit{bad, poor, fair, good, excellent}\} denotes the established textual rating categories as defined by the ITU. We convert the videos into sequences of keyframes, which are sampled as the first frame of every second. 
Then we form the question-answer pairs like `\textit{How would you rate the quality of the video? $|keyframe1| |keyframe2|$ ... The quality of the video is bad/poor/fair/good/excellent}' to fine-tune the LMM. 

After training, we can prompt LMM with the same question-answer structure and obtain the responded \textit{[SCORE\_TOKEN]} from the `\textit{The quality of the video is [SCORE\_TOKEN]}'. The \textit{[SCORE\_TOKEN]} can then be converted to the log probabilities of \{\textit{bad, poor, fair, good, excellent}\}. Finally, we conduct a close-set softmax on log probabilities to get the probabilities $p_{l_i}$ for each level ($p_{l_i}$ for all $l_i$ sum as 1):

\begin{equation}
    p_{l_i} = \frac{e^{\mathcal{X}_{l_i}}}{\sum_{j=1}^{5} {e^{\mathcal{X}_{l_j}}}}
\end{equation}
and the final predicted scores of LMMs can be derived as 
\begin{equation}
    \mathrm{S_{LMM}}=\sum_{i=1}^5 p_{l_i} G(l_i) = i \times  \frac{e^{\mathcal{X}_{l_i}}}{\sum_{j=1}^{5} {e^{\mathcal{X}_{l_j}}}}
    \label{eq:5}
\end{equation}

During the efficiency test, we find the Q-Align takes up about 8,179M parameters and 991G MACs. Q-Align deals with every 30fps video clip for about 533ms on GPU 3090.

\paragraph{Implementation details}

We use the PyTorch framework.
In experiments, we set batch sizes as 64 and the learning rate is set as $2e-5$. We select mPLUG-Owl-2 as the LMM model. We only train the model on the training set of YT-UGC. We train for 2 epochs for all variants, which takes up about 50 minutes. We conduct training on 4*NVIDIA A100 80G GPUs, and report inference latency on one RTX3090 24G GPU. For videos, we sample at rate \textit{1fps}. The sampled frames are padded to square and then resized to $448\times448$.

\begin{figure*}[!ht]
    \centering
    \includegraphics[width=0.95\linewidth]{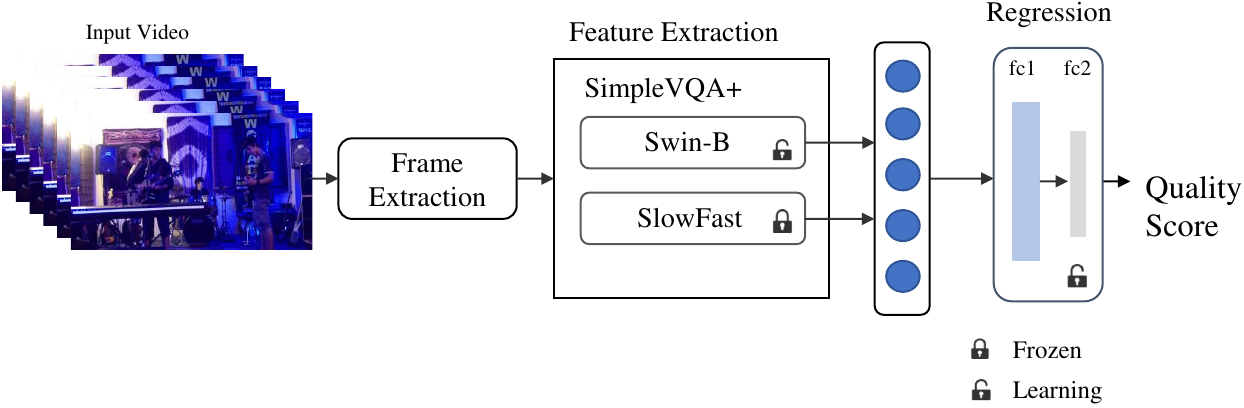}
    \caption{The framework of SimpleVQA+ \cite{sun2022deep, sun2023analysis} proposed by Team SJTU MMLab.}
    \label{fig:team4}
\end{figure*}

\subsection{Blind Video Quality Assessment Models through Spatial and Temporal Quality-Aware Features}


\begin{center}

\vspace{2mm}
\noindent\emph{\textbf{Team SJTU MMLab}}
\vspace{2mm}

\noindent\emph{Wei Sun,
Yuqin Cao,
Yanwei Jiang,
Jun Jia,
Zhichao Zhang,
Zijian Chen,
Weixia Zhang,
Xiongkuo Min}

\vspace{2mm}

\noindent\emph{Shanghai Jiao Tong University}

\vspace{2mm}

\noindent{\emph{Contact: \url{suguwei@sjtu.edu.cn}}}

\end{center}

The proposed BVQA model is based on SimpleVQA+ \cite{sun2022deep, sun2023analysis}, comprising the Swin Transformer-B~\cite{liu2021swin} for spatial feature extraction from key frames, and a temporal pathway of SlowFast for temporal feature extraction from video chunks. Then, we concatenate these features and fuse them into the final quality score via a two-layer MLP. The model is shown in \cref{fig:team4}.

We trained SimpleVQA+ on the LSVQ dataset\cite{ying2020live}. We utilize LSVQ \cite{ying2020live} and YT-UGC dataset \cite{wang2021rich} for training. During the pre-processing process, we sample one key frame from one-second video chunks (i.e. $1$ fps) for the spatial feature extraction module. The resolution of key frames is further resized to $384\times 384$ for training. For the temporal feature extraction module, the resolution of the videos is resized to $224\times 224$. We then split the whole video into several one-second length video chunks to extract the corresponding temporal features. 

We train the proposed model on $2$ Nvidia RTX 3090 GPUs with a batch size $6$ for $30$ epochs ($\approx3$hrs). The learning rate is set as $10^{-5}$. During the inference phase, we feed the video into two models which are trained on the LSVQ and YT-UGC datasets respectively, to obtain prediction scores. Then, we average two scores to obtain the final prediction score. Our proposed model is trained efficiently and can take advantage of other quality-aware pre-trained features, which can help decrease the risk of overfitting.

\begin{figure*}[t]
    \centering
    \includegraphics{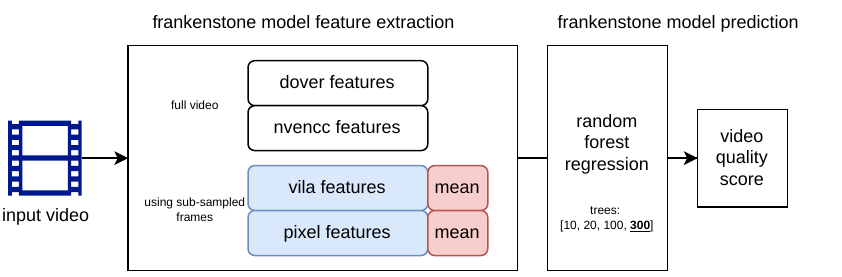}
    \caption{Overview of the \texttt{frankenstone} model proposed by Team AVT.}
    \label{fig:avt}
\end{figure*}

\subsection{Frankenstone -- a video quality prediction model combining other models and features}
\label{sec:AVT}


\begin{center}

\vspace{2mm}
\noindent\emph{\textbf{Team AVT}}
\vspace{2mm}

\noindent\emph{Steve Göring}

\vspace{2mm}

\noindent\emph{Audiovisual Technology Group; Technische Universität Ilmenau; Germany}

\vspace{2mm}

\noindent{\emph{Contact: \url{steve.goering@tu-ilmenau.de}}}

\end{center}

The Frankenstone model uses several other models/features as a baseline and combines them with a random forest regression, similar to \cite{goering2021pixel}.
Four main groups are used as features, for each feature value mean aggregation is performed. 
For example, NVENCC is used to extract meta-data and encoding properties (such as bitrate for a specific encoding setting).
Furthermore, the DOVER model~\cite{wu2023dover}\footnote{\url{https://github.com/VQAssessment/DOVER}} score and two of its atomic features are used in the Frankenstone model.

In addition, signal-based features, e.g. SI, TI, colorfulness, average luminance, for a subset of the frames are extracted, and on the same subset also VILA model~\cite{ke2023vila}\footnote{\url{https://github.com/google-research/google-research/tree/master/vila}} predictions (image appeal) are performed.

The subset of processed frames is done in two steps, the first samples for each second of the video the first frame.
The second step takes the sampled frames and reduces them with more importance to the end of the video.
That means for a 20\,s 30 fps video, 20 frames are sampled, and then out of them, the following 5 frames are used: [0, 6, 11, 15, 18].

All features are extracted in separate threads to make the model faster. Afterwards, the Frankenstone model combines the mentioned features and scores using a Random Forest Regression model. AVT uses DOVER~\cite{wu2023dover} for user-generated video quality prediction, and VILA for per-frame image appeal~\cite{ke2023vila} prediction. Only the YouTube UGC~\cite{wang2019youtube} training data was used.


In Fig.~\ref{fig:avt} an overview of the model structure is provided. The video is fed into the model and then several features are calculated in threads (parallel computation), dover\footnote{dover does also frame sub-sampling} and nvencc features (height, aspect ratio, bitrate for a specific encoding) are calculated for the full video, while pixel (SI, TI, colorfulness, average luminance, sharpness, nima appeal/quality~\cite{idealods2018imagequalityassessment}, TI calculations to the first frame, SSIM pairwise and to the first frame) and vila features are only calculated for a subset of the video frames (because otherwise, the model would not hit the runtime requirements).
The extracted features are combined using a random forest regression (during model development with a varying number of trees, the submitted model uses 300 trees).

The runtime of the model has been evaluated exemplarily with various videos, in the following the \texttt{Sports\_2160P-210c.mkv} (30 fps, UHD-1, 20s duration) video is used.
The 24 time measurements result in an average runtime of $\approx19.616\,s$, with a standard derivation of $\approx0.138\,s$.
However, this may vary, depending on a warm start of the model (and corresponding file-system caches). The model may not be fast enough for smaller videos, because the data must be transferred to the GPU first.


\paragraph{Implementation details}

\begin{itemize}
    \item \textbf{Framework:} For feature extraction mainly Tensorflow is used, however, some of the included models rely on PyTorch, and the final score is predicted with a random forest regression model (part of Tensorflow Decision Forests package).
    \item \textbf{Optimizer and Learning Rate:} A random forest model with a variable number of trees (10, 20, 100, and 300) have been used, there was no improvement using more trees, the final model has 300 trees.
    \item \textbf{GPU:} NVIDIA GeForce RTX 3090 Ti (24 GB)
    \item \textbf{Datasets:} Youtube UGC training data, no augmentation.
    \item \textbf{Training Time:} Extraction of features for each video $\approx$ 20\,s max, thus 892 training videos, $\approx$ 12\,h extraction time (was performed with 3-4 parallel processes to reduce the time, overall on one PC), training the random forest regression model takes $<1$\,min (part of the Tensorflow Decision Forests package).
    \item \textbf{Efficiency Optimization Strategies:} Performing feature extraction in parallel threads.
\end{itemize}

\subsection{Ranking-based training strategy in siamese manner}
\label{sec:bvi}


\begin{center}

\vspace{2mm}
\noindent\emph{\textbf{Team BVI-VQA}}
\vspace{2mm}

\noindent\emph{Zihao Qi, Chen Feng}

\vspace{2mm}

\noindent\emph{Visual Information Laboratory, University of Bristol}

\vspace{2mm}

\noindent{\emph{Contact: \url{zihao.qi@bristol.ac.uk}}}

\end{center}

The team uses FasterVQA~\cite{wu2023neighbourhood} as backbone, training in a siamese manner. During training, the siamese network takes a pair of videos as input and tries to predict which one is in better quality. This training strategy, following a similar methodology proposed in previous works~\cite{feng2022rankdvqa,qi2023full}, makes it possible to train our model on multiple datasets with various scoring scale (YouTube-UGC~\cite{wang2019youtube}, LIVE-VQC~\cite{sinno2018large}, KoNVid-1k). After trained in siamese manner, the FasterVQA model is then fine-tuned on YouTube-UGC.

Based on the intuition to train our model over multiple datasets, we proposed a ranking-based training strategy to train an existing SOTA network, FasterVQA~\cite{wu2023neighbourhood}, in a siamese manner.

\begin{figure}[t]
    \centering
    \begin{minipage}[b]{\linewidth}
      \centering
      \centerline{\includegraphics[width=\textwidth]{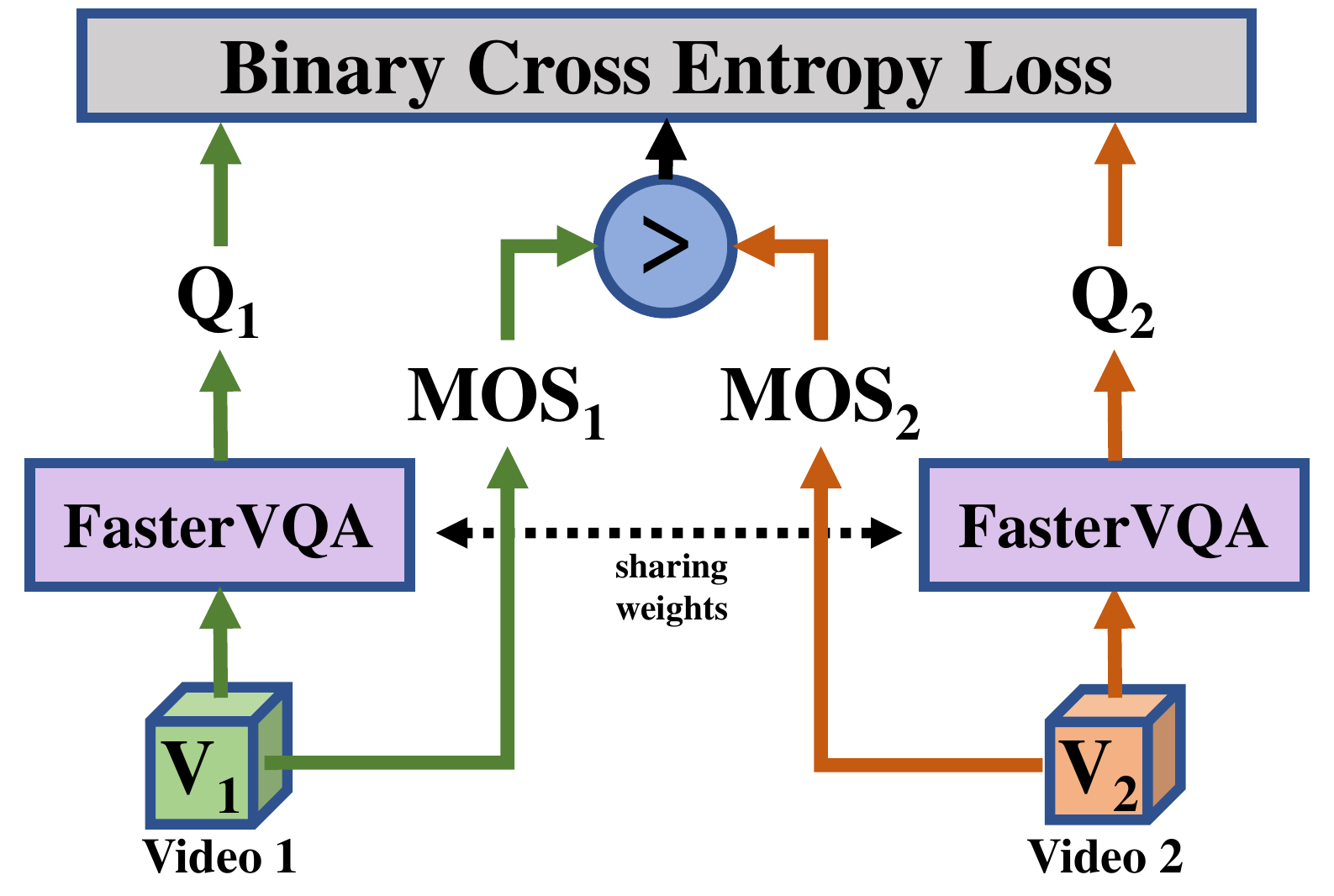}}
    \end{minipage}
    \caption{Overall picture of the siamese training process proposed by Team BVI-VQA.}
    \label{fig:bristol}
\end{figure}

A common challenge when training on multiple datasets is: different datasets usually have inconsistent scoring scale and crowdsourcing protocol. To solve this problem, we trained our model using a siamese structure, consisting of two FasterVQA networks sharing the same weights. At each time, the siamese network takes a random pair of videos from the same dataset as input and learns to predict which one is of the better quality (with higher MOS ground-truth value). Because the network does not directly take MOS as training labels, it avoids the problem that MOS from different datasets may have different scoring scale. This ranking-based training strategy shares a similar insight as previous works~\cite{feng2022rankdvqa,qi2023full}. Pre-trained model from FasterVQA has been used to initialize the training. After trained 20 epoches over 3 datasets (YouTube-UGC~\cite{wang2019youtube}, LIVE-VQC~\cite{sinno2018large}, KoNVid-1k) in siamese manner, the model is then finetuned on YouTube-UGC. The training framework is illustrated in \cref{fig:bristol}.

\begin{table}[t]
    \centering
    \begin{tabular}{l c}
         \toprule
         Method & SROC  \\
         \midrule
         \textbf{FasterVQA with Siamese Training} & 0.818  \\
         Pre-trained FasterVQA & 0.813  \\
         Pre-trained SimpleVQA & 0.792  \\
         \bottomrule
    \end{tabular}
    \caption{Ablation study on the testing set by Team BVI-VQA.}
    \label{tab:bvi-ablation}
\end{table}

\paragraph{Implementation details}

\begin{itemize}
    \item \textbf{Framework:} PyTorch.
    \item \textbf{Optimizer and Learning Rate:} AdamW using learning rate 1e-4 and weight decay 0.05.
    \item \textbf{GPU:} NVIDIA RTX 3090.
    \item \textbf{Datasets:} YouTube-UGC, LIVE-VQC, KoNVid-1k.
    \item \textbf{Training Time:} 12h.
\end{itemize}

\section*{Acknowledgements}

This work was partially supported by the Humboldt Foundation. We thank the AIS 2024 sponsors: Meta Reality Labs, Meta, Netflix, Sony Interactive Entertainment (FTG), and the University of W\"urzburg (Computer Vision Lab).

The challenge organizers thank Ioannis Katsavounidis (Meta), Christos Bampis (Netflix), and Balu Adsumilli (Google) for their feedback.

{\small
\bibliographystyle{ieeenat_fullname}
\bibliography{refs}

\begin{thebibliography}{45}
\providecommand{\natexlab}[1]{#1}
\providecommand{\url}[1]{\texttt{#1}}
\expandafter\ifx\csname urlstyle\endcsname\relax
  \providecommand{\doi}[1]{doi: #1}\else
  \providecommand{\doi}{doi: \begingroup \urlstyle{rm}\Url}\fi

\bibitem[Barman and Martini(2019)]{Barman2019Survey}
Nabajeet Barman and Maria~G. Martini.
\newblock {QoE Modeling for HTTP Adaptive Video Streaming–A Survey and Open Challenges}.
\newblock \emph{IEEE Access}, 7:\penalty0 30831--30859, 2019.

\bibitem[Conde et~al.(2024{\natexlab{a}})Conde, Lei, Li, Katsavounidis, Timofte, et~al.]{conde2024ais_sr}
Marcos~V. Conde, Zhijun Lei, Wen Li, Ioannis Katsavounidis, Radu Timofte, et~al.
\newblock Real-time 4k super-resolution of compressed {AVIF} images. {AIS} 2024 challenge survey.
\newblock In \emph{Proceedings of the IEEE/CVF Conference on Computer Vision and Pattern Recognition Workshops}, 2024{\natexlab{a}}.

\bibitem[Conde et~al.(2024{\natexlab{b}})Conde, Zadtootaghaj, Barman, Timofte, et~al.]{conde2024ais_vqa}
Marcos~V. Conde, Saman Zadtootaghaj, Nabajeet Barman, Radu Timofte, et~al.
\newblock {AIS} 2024 challenge on video quality assessment of user-generated content: Methods and results.
\newblock In \emph{Proceedings of the IEEE/CVF Conference on Computer Vision and Pattern Recognition Workshops}, 2024{\natexlab{b}}.

\bibitem[Feng et~al.(2022)Feng, Danier, Zhang, and Bull]{feng2022rankdvqa}
Chen Feng, Duolikun Danier, Fan Zhang, and David~R Bull.
\newblock {RankDVQA}: Deep vqa based on ranking-inspired hybrid training.
\newblock \emph{arXiv preprint arXiv:2202.08595}, 2022.

\bibitem[Ghadiyaram(2017)]{FRIQUEE}
Deepti Ghadiyaram.
\newblock Perceptual quality prediction on authentically distorted images using a bag of features approach.
\newblock \emph{Journal of Vision}, 17(1)\penalty0 (32):\penalty0 1--25, 2017.

\bibitem[G\"oring et~al.(2021)G\"oring, {Rao Ramachandra Rao}, Feiten, and Raake]{goering2021pixel}
Steve G\"oring, Rakesh {Rao Ramachandra Rao}, Bernhard Feiten, and Alexander Raake.
\newblock Modular framework and instances of pixel-based video quality models for uhd-1/4k.
\newblock \emph{IEEE Access}, 9:\penalty0 31842--31864, 2021.

\bibitem[G\"otz-Hahn et~al.(2021)G\"otz-Hahn, Hosu, Lin, and Saupe]{hahn2021}
Franz G\"otz-Hahn, Vlad Hosu, Hanhe Lin, and Dietmar Saupe.
\newblock {KonVid-150k: A Dataset for No-Reference Video Quality Assessment of Videos in-the-Wild}.
\newblock In \emph{IEEE Access 9}, pages 72139--72160. IEEE, 2021.

\bibitem[Gu et~al.(2022)Gu, Cai, Dong, Ren, Timofte, Gong, Lao, Shi, Wang, Yang, et~al.]{gu2022ntire}
Jinjin Gu, Haoming Cai, Chao Dong, Jimmy~S Ren, Radu Timofte, Yuan Gong, Shanshan Lao, Shuwei Shi, Jiahao Wang, Sidi Yang, et~al.
\newblock Ntire 2022 challenge on perceptual image quality assessment.
\newblock In \emph{Proceedings of the IEEE/CVF conference on computer vision and pattern recognition}, pages 951--967, 2022.

\bibitem[He et~al.(2024)He, He, Zhu, Zeng, Fan, and Tu]{He2024cover}
Chenlong He, Chenlong He, Ruoxi Zhu, Xiaoyang Zeng, Yibo Fan, and Zhengzhong Tu.
\newblock {COVER}: A comprehensive video quality evaluator.
\newblock In \emph{Proceedings of the IEEE/CVF Conference on Computer Vision and Pattern Recognition Workshops}, 2024.

\bibitem[Hosu et~al.(2017)Hosu, Hahn, Jenadeleh, Lin, Men, Szir{\'a}nyi, Li, and Saupe]{hosu2017konstanz}
Vlad Hosu, Franz Hahn, Mohsen Jenadeleh, Hanhe Lin, Hui Men, Tam{\'a}s Szir{\'a}nyi, Shujun Li, and Dietmar Saupe.
\newblock The konstanz natural video database (konvid-1k).
\newblock In \emph{2017 Ninth international conference on quality of multimedia experience (QoMEX)}, pages 1--6. IEEE, 2017.

\bibitem[Ke et~al.(2023)Ke, Ye, Yu, Wu, Milanfar, and Yang]{ke2023vila}
Junjie Ke, Keren Ye, Jiahui Yu, Yonghui Wu, Peyman Milanfar, and Feng Yang.
\newblock Vila: Learning image aesthetics from user comments with vision-language pretraining.
\newblock In \emph{Proceedings of the IEEE/CVF Conference on Computer Vision and Pattern Recognition}, pages 10041--10051, 2023.

\bibitem[Korhonen(2019)]{korhonen2019two}
Jari Korhonen.
\newblock Two-level approach for no-reference consumer video quality assessment.
\newblock \emph{IEEE Trans. Image Process.}, 28\penalty0 (12):\penalty0 5923--5938, 2019.

\bibitem[Kundu et~al.(2017)Kundu, Ghadiyaram, Bovik, and Evans]{kundu2017no}
Debarati Kundu, Deepti Ghadiyaram, Alan~C Bovik, and Brian~L Evans.
\newblock No-reference quality assessment of tone-mapped {HDR} pictures.
\newblock \emph{IEEE Trans. Image Process.}, 26\penalty0 (6):\penalty0 2957--2971, 2017.

\bibitem[Lennan et~al.(2018)Lennan, Nguyen, and Tran]{idealods2018imagequalityassessment}
Christopher Lennan, Hao Nguyen, and Dat Tran.
\newblock Image quality assessment.
\newblock \url{https://github.com/idealo/image-quality-assessment}, 2018.

\bibitem[Liu et~al.(2021)Liu, Lin, Cao, Hu, Wei, Zhang, Lin, and Guo]{liu2021swin}
Ze Liu, Yutong Lin, Yue Cao, Han Hu, Yixuan Wei, Zheng Zhang, Stephen Lin, and Baining Guo.
\newblock Swin transformer: Hierarchical vision transformer using shifted windows.
\newblock In \emph{Proceedings of the IEEE/CVF international conference on computer vision}, pages 10012--10022, 2021.

\bibitem[Liu et~al.(2022)Liu, Mao, Wu, Feichtenhofer, Darrell, and Xie]{liu2022convnet}
Zhuang Liu, Hanzi Mao, Chao-Yuan Wu, Christoph Feichtenhofer, Trevor Darrell, and Saining Xie.
\newblock A convnet for the 2020s.
\newblock In \emph{Proceedings of the IEEE/CVF conference on computer vision and pattern recognition}, pages 11976--11986, 2022.

\bibitem[Mittal et~al.(2012)Mittal, Moorthy, and Bovik]{mittal2012no}
Anish Mittal, Anush~Krishna Moorthy, and Alan~Conrad Bovik.
\newblock No-reference image quality assessment in the spatial domain.
\newblock \emph{IEEE Trans. Image Process.}, 21\penalty0 (12):\penalty0 4695--4708, 2012.

\bibitem[Mittal et~al.(2013)Mittal, Soundararajan, and Bovik]{6353522}
Anish Mittal, Rajiv Soundararajan, and Alan~C. Bovik.
\newblock Making a “completely blind” image quality analyzer.
\newblock \emph{IEEE Signal Processing Letters}, 20\penalty0 (3):\penalty0 209--212, 2013.

\bibitem[Netflix()]{vmaf}
Netflix.
\newblock {VMAF - Video Multi-Method Assessment Fusion}.
\newblock \url{https://github.com/Netflix/vmaf}.

\bibitem[Qi et~al.(2023)Qi, Feng, Danier, Zhang, Xu, Liu, and Bull]{qi2023full}
Zihao Qi, Chen Feng, Duolikun Danier, Fan Zhang, Xiaozhong Xu, Shan Liu, and David Bull.
\newblock Full-reference video quality assessment for user generated content transcoding.
\newblock \emph{arXiv preprint arXiv:2312.12317}, 2023.

\bibitem[Radford et~al.(2021)Radford, Kim, Hallacy, Ramesh, Goh, Agarwal, Sastry, Askell, Mishkin, Clark, et~al.]{radford2021learning}
Alec Radford, Jong~Wook Kim, Chris Hallacy, Aditya Ramesh, Gabriel Goh, Sandhini Agarwal, Girish Sastry, Amanda Askell, Pamela Mishkin, Jack Clark, et~al.
\newblock Learning transferable visual models from natural language supervision.
\newblock In \emph{International conference on machine learning}, pages 8748--8763. PMLR, 2021.

\bibitem[Sandler et~al.(2018)Sandler, Howard, Zhu, Zhmoginov, and Chen]{sandler2018mobilenetv2}
Mark Sandler, Andrew Howard, Menglong Zhu, Andrey Zhmoginov, and Liang-Chieh Chen.
\newblock Mobilenetv2: Inverted residuals and linear bottlenecks.
\newblock In \emph{Proceedings of the IEEE conference on computer vision and pattern recognition}, pages 4510--4520, 2018.

\bibitem[Sinno and Bovik(2018)]{sinno2018large}
Zeina Sinno and Alan~Conrad Bovik.
\newblock Large-scale study of perceptual video quality.
\newblock \emph{IEEE Transactions on Image Processing}, 28\penalty0 (2):\penalty0 612--627, 2018.

\bibitem[Statista({\natexlab{a}})]{statOTT}
Statista.
\newblock {Number of users of OTT video worldwide from 2020 to 2029 (in millions) [Graph]}.
\newblock \url{https://www.statista.com/forecasts/1207843/ott-video-users-worldwide/}, {\natexlab{a}}.

\bibitem[Statista({\natexlab{b}})]{statpermin}
Statista.
\newblock {Daily time spent on social networking by internet users worldwide from 2012 to 2024 (in minutes) [Graph]}.
\newblock \url{https://www.statista.com/statistics/433871/daily-social-media-usage-worldwide/}, {\natexlab{b}}.

\bibitem[Sun et~al.(2022)Sun, Min, Lu, and Zhai]{sun2022deep}
Wei Sun, Xiongkuo Min, Wei Lu, and Guangtao Zhai.
\newblock A deep learning based no-reference quality assessment model for ugc videos.
\newblock In \emph{Proceedings of the 30th ACM International Conference on Multimedia}, pages 856--865, 2022.

\bibitem[Sun et~al.(2023)Sun, Wen, Min, Lan, Zhai, and Ma]{sun2023analysis}
Wei Sun, Wen Wen, Xiongkuo Min, Long Lan, Guangtao Zhai, and Kede Ma.
\newblock Analysis of video quality datasets via design of minimalistic video quality models.
\newblock \emph{arXiv preprint arXiv:2307.13981}, 2023.

\bibitem[Tu et~al.(2021{\natexlab{a}})Tu, Wang, Birkbeck, Adsumilli, and Bovik]{9405420}
Zhengzhong Tu, Yilin Wang, Neil Birkbeck, Balu Adsumilli, and Alan~C. Bovik.
\newblock {UGC-VQA}: Benchmarking blind video quality assessment for user generated content.
\newblock \emph{IEEE Trans. Image Process.}, 30:\penalty0 4449--4464, 2021{\natexlab{a}}.

\bibitem[Tu et~al.(2021{\natexlab{b}})Tu, Yu, Wang, Birkbeck, Adsumilli, and Bovik]{tu2021rapique}
Zhengzhong Tu, Xiangxu Yu, Yilin Wang, Neil Birkbeck, Balu Adsumilli, and Alan~C Bovik.
\newblock {RAPIQUE}: {R}apid and accurate video quality prediction of user generated content.
\newblock \emph{IEEE Open Journal of Signal Processing}, 2:\penalty0 425--440, 2021{\natexlab{b}}.

\bibitem[Tu et~al.(2022)Tu, Talebi, Zhang, Yang, Milanfar, Bovik, and Li]{tu2022maxim}
Zhengzhong Tu, Hossein Talebi, Han Zhang, Feng Yang, Peyman Milanfar, Alan Bovik, and Yinxiao Li.
\newblock Maxim: Multi-axis mlp for image processing.
\newblock In \emph{Proceedings of the IEEE/CVF conference on computer vision and pattern recognition}, pages 5769--5780, 2022.

\bibitem[Utke et~al.(2022)Utke, Zadtootaghaj, Schmidt, Bosse, and M{\"o}ller]{ndnetgaming}
Markus Utke, Saman Zadtootaghaj, Steven Schmidt, Sebastian Bosse, and Sebastian M{\"o}ller.
\newblock Ndnetgaming-development of a no-reference deep cnn for gaming video quality prediction.
\newblock \emph{Multimedia Tools and Applications}, pages 1--23, 2022.

\bibitem[Wang et~al.(2023)Wang, Chan, and Loy]{wang2023exploring}
Jianyi Wang, Kelvin~CK Chan, and Chen~Change Loy.
\newblock Exploring clip for assessing the look and feel of images.
\newblock In \emph{Proceedings of the AAAI Conference on Artificial Intelligence}, pages 2555--2563, 2023.

\bibitem[Wang et~al.(2019)Wang, Inguva, and Adsumilli]{wang2019youtube}
Yilin Wang, Sasi Inguva, and Balu Adsumilli.
\newblock Youtube ugc dataset for video compression research.
\newblock In \emph{2019 IEEE 21st International Workshop on Multimedia Signal Processing (MMSP)}, pages 1--5. IEEE, 2019.

\bibitem[Wang et~al.(2021)Wang, Ke, Talebi, Yim, Birkbeck, Adsumilli, Milanfar, and Yang]{wang2021rich}
Yilin Wang, Junjie Ke, Hossein Talebi, Joong~Gon Yim, Neil Birkbeck, Balu Adsumilli, Peyman Milanfar, and Feng Yang.
\newblock Rich features for perceptual quality assessment of ugc videos.
\newblock In \emph{Proceedings of the IEEE/CVF Conference on Computer Vision and Pattern Recognition}, pages 13435--13444, 2021.

\bibitem[Wang et~al.(2024)Wang, Gao, Wu, Conde, Timofte, Liu, Chen, et~al.]{wang2024ais_event}
Zuowen Wang, Chang Gao, Zongwei Wu, Marcos~V. Conde, Radu Timofte, Shih-Chii Liu, Qinyu Chen, et~al.
\newblock {E}vent-{B}ased {E}ye {T}racking. {AIS} 2024 {C}hallenge {S}urvey.
\newblock In \emph{Proceedings of the IEEE/CVF Conference on Computer Vision and Pattern Recognition Workshops}, 2024.

\bibitem[Wen et~al.(2024)Wen, Li, Zhang, Liao, Li, Zhang, and Ma]{wen2024modular}
Wen Wen, Mu Li, Yabin Zhang, Yiting Liao, Junlin Li, Li Zhang, and Kede Ma.
\newblock {Modular Blind Video Quality Assessment}, 2024.

\bibitem[Wu et~al.(2022)Wu, Chen, Hou, Liao, Wang, Sun, Yan, and Lin]{wu2022fast}
Haoning Wu, Chaofeng Chen, Jingwen Hou, Liang Liao, Annan Wang, Wenxiu Sun, Qiong Yan, and Weisi Lin.
\newblock Fast-vqa: Efficient end-to-end video quality assessment with fragment sampling.
\newblock In \emph{European conference on computer vision}, pages 538--554. Springer, 2022.

\bibitem[Wu et~al.(2023{\natexlab{a}})Wu, Chen, Liao, Hou, Sun, Yan, Gu, and Lin]{wu2023neighbourhood}
Haoning Wu, Chaofeng Chen, Liang Liao, Jingwen Hou, Wenxiu Sun, Qiong Yan, Jinwei Gu, and Weisi Lin.
\newblock Neighbourhood representative sampling for efficient end-to-end video quality assessment.
\newblock \emph{IEEE Transactions on Pattern Analysis and Machine Intelligence}, 2023{\natexlab{a}}.

\bibitem[Wu et~al.(2023{\natexlab{b}})Wu, Zhang, Liao, Chen, Hou, Wang, Sun, Yan, and Lin]{wu2023dover}
Haoning Wu, Erli Zhang, Liang Liao, Chaofeng Chen, Jingwen~Hou Hou, Annan Wang, Wenxiu~Sun Sun, Qiong Yan, and Weisi Lin.
\newblock Exploring video quality assessment on user generated contents from aesthetic and technical perspectives.
\newblock In \emph{International Conference on Computer Vision (ICCV)}, 2023{\natexlab{b}}.

\bibitem[Wu et~al.(2023{\natexlab{c}})Wu, Zhang, Zhang, Chen, Liao, Li, Gao, Wang, Zhang, Sun, et~al.]{wu2023q}
Haoning Wu, Zicheng Zhang, Weixia Zhang, Chaofeng Chen, Liang Liao, Chunyi Li, Yixuan Gao, Annan Wang, Erli Zhang, Wenxiu Sun, et~al.
\newblock Q-align: Teaching lmms for visual scoring via discrete text-defined levels.
\newblock \emph{arXiv preprint arXiv:2312.17090}, 2023{\natexlab{c}}.

\bibitem[Xue et~al.(2014)Xue, Mou, Zhang, Bovik, and Feng]{xue2014blind}
Wufeng Xue, Xuanqin Mou, Lei Zhang, Alan~C Bovik, and Xiangchu Feng.
\newblock Blind image quality assessment using joint statistics of gradient magnitude and laplacian features.
\newblock \emph{IEEE Trans. Image Process.}, 23\penalty0 (11):\penalty0 4850--4862, 2014.

\bibitem[Ye et~al.(2012)Ye, Kumar, Kang, and Doermann]{6247789}
Peng Ye, Jayant Kumar, Le Kang, and David Doermann.
\newblock Unsupervised feature learning framework for no-reference image quality assessment.
\newblock In \emph{Proc. IEEE Conf. Comput. Vis. Pattern Recognit. (CVPR)}, pages 1098--1105, 2012.

\bibitem[Ying et~al.(2020)Ying, Mandal, Ghadiyaram, and Bovik]{ying2020live}
Z Ying, M Mandal, D Ghadiyaram, and AC Bovik.
\newblock Live large-scale social video quality (lsvq) database.
\newblock \emph{Online: https://github. com/baidut/PatchVQ}, 2020.

\bibitem[Ying et~al.(2021)Ying, Mandal, Ghadiyaram, and Bovik]{ying2021patch}
Zhenqiang Ying, Maniratnam Mandal, Deepti Ghadiyaram, and Alan Bovik.
\newblock Patch-vq:'patching up'the video quality problem.
\newblock In \emph{Proceedings of the IEEE/CVF conference on computer vision and pattern recognition}, pages 14019--14029, 2021.

\bibitem[Zheng et~al.(2024)Zheng, Tu, Madhusudana, Zeng, Bovik, and Fan]{zheng2024faver}
Qi Zheng, Zhengzhong Tu, Pavan~C Madhusudana, Xiaoyang Zeng, Alan~C Bovik, and Yibo Fan.
\newblock Faver: Blind quality prediction of variable frame rate videos.
\newblock \emph{Signal Processing: Image Communication}, 122:\penalty0 117101, 2024.

\end{thebibliography}
}

\end{document}